\documentclass[sigconf]{acmart}

\usepackage{microtype}
\usepackage{subcaption}
\usepackage{tabularx}
    \newcolumntype{Y}{>{\raggedright\arraybackslash}X}
\AtBeginDocument{%
  }

\copyrightyear{2024}
\acmYear{2024}
\setcopyright{acmlicensed}\acmConference[GoodIT '24]{International
Conference on Information Technology for Social Good}{September 4--6,
2024}{Bremen, Germany}
\acmBooktitle{International Conference on Information Technology for Social
Good (GoodIT '24), September 4--6, 2024, Bremen, Germany}
\acmDOI{10.1145/3677525.3678694}
\acmISBN{979-8-4007-1094-0/24/09}

\begin{document}

\title{MoralBERT: A Fine-Tuned Language Model for Capturing Moral Values in Social Discussions}

\author{Vjosa Preniqi}

\orcid{1234-5678-9012}
\authornotemark[1]
\affiliation{%
  \institution{Queen Mary University of London}
  \city{London}
  \country{UK}}
\email{v.preniqi@qmul.ac.uk}

\author{Iacopo Ghinassi}
\affiliation{%
  \institution{Queen Mary University of London}
  \city{London}
  \country{UK}}
\email{i.ghinassi@qmul.ac.uk}

\author{Julia Ive}
\affiliation{%
  \institution{Queen Mary University of London}
  \city{London}
  \country{UK}}
\email{j.ive@qmul.ac.uk}

\author{Charalampos Saitis}
\affiliation{%
  \institution{Queen Mary University of London}
  \city{London}
  % \state{Beijing Shi}
  \country{UK}}
\email{c.saitis@qmul.ac.uk}

\author{Kyriaki Kalimeri}
\affiliation{%
 \institution{ISI Foundation}
 \city{Turin}
 \country{Italy}}
\email{kyriaki.kalimeri@isi.it}

\renewcommand{\shortauthors}{Preniqi et al.}

\begin{abstract}
Moral values play a fundamental role in how we evaluate information, make decisions, and form judgements around important social issues.
Controversial topics, including vaccination, abortion, racism, and sexual orientation, often elicit opinions and attitudes that are not solely based on evidence but rather reflect moral worldviews.  
Recent advances in Natural Language Processing (NLP) show that moral values can be gauged in human-generated textual content. Building on the Moral Foundations Theory (MFT), this paper introduces MoralBERT, a range of language representation models fine-tuned to capture moral sentiment in social discourse. 
We describe a framework for both aggregated and domain-adversarial training on multiple heterogeneous MFT human-annotated datasets sourced from Twitter (now X), Reddit, and Facebook that broaden textual content diversity in terms of social media audience interests, content presentation and style, and spreading patterns. 
We show that the proposed framework achieves an average F1 score that is between 11\% and 32\% higher than lexicon-based approaches, Word2Vec embeddings, and zero-shot classification with large language models such as GPT-4 for in-domain inference. Domain-adversarial training yields better out-of domain predictions than aggregate training while achieving comparable performance to zero-shot learning. 
Our approach contributes to annotation-free and effective morality learning, and provides useful insights towards a more comprehensive understanding of moral narratives in controversial social debates using NLP.
\end{abstract}




\maketitle

\section{Introduction}
Language is not merely a tool for communication, but a reflection of a plethora of intricate psychological constructs. The words and phrases people use can reveal underlying emotions~\cite{nandwani2021review}, personality traits~\cite{schwartz2013personality}, and even moral values~\cite{Graham2013}. The latter occupy a salient position, significantly influencing stance taking on contentious social issues such as vaccine hesitancy \cite{kalimeri2019human}, civil unrest \cite{Mooijman2018}, but also personal taste such as the type of music we like to listen~\cite{preniqi2023soundscapes,preniqi2022more}.
Here, we aim to improve the automatic assessment of moral values in text. This is an important task considering that a comprehensive understanding of moral values at a broader scale could greatly contribute to timely insights into attitudes and judgments concerning social issues, mitigating social polarisation or even uprisings through enhancing the efficacy of communication campaigns~\cite{Kalimeri2019}. 

We employ the Moral Foundations Theory (MFT) as the theoretical underpinning to operationalise morality in the following six psychological ``foundations'' of moral reasoning, divided into ``virtue/vice'' polarities \cite{haidt2007morality, haidt2004intuitive,haidt2012righteous}: \textit{Care/Harm} involves concern for others' suffering and includes virtues like empathy and compassion; \textit{Fairness/Cheating} focuses on issues of unfair treatment, inequality, and justice; \textit{Loyalty/Betrayal} pertains to group obligations such as loyalty and the vigilance against betrayal; \textit{Authority/Subversion} centers on social order and hierarchical responsibilities, emphasising obedience and respect; \textit{Purity/Degradation} relates to physical and spiritual sanctity, incorporating virtues like chastity and self-control;  \textit{Liberty/Oppression} addresses feelings of reactance and resentment towards oppressors. 

Alongside MFT came a lexicon to guide morality detection in text, illustrating the importance of and need for studying human morality as it manifests in verbal expression, but also highlighting the challenges of the task~\cite{graham2009liberals}.
As interest in language and morality has grown, improved dictionaries and other Natural Language Processing (NLP) resources have been developed to study the role of moral values in human life, including ground truth datasets with moral annotations~\cite{hopp2021extended,Araque2020,guo2023data,nguyen2024measuring,hoover2020moral,trager2022moral,beiro2023moral}.
These works offer quantitative evidence that people project their moral worldviews in a variety of social topics, from the emergence of symbolism and aesthetics of the
resistance movement~\cite{mejova2023authority} to public perceptions during the COVID-19 vaccination campaigns~\cite{borghouts2023understanding,beiro2023moral}, amongst others. 

Also building on the MFT, here we introduce MoralBERT, a range of language representation models fine-tuned to capture moral sentiment in social discourse. 
We describe a framework for both aggregated and domain-adversarial training on multiple heterogeneous MFT human-annotated datasets sourced from Twitter (before it was rebranded as X), Reddit, and Facebook that broaden textual content diversity in terms of social media audience interests, content presentation and style, and spreading patterns. 
We show that the proposed framework achieves an average F1 score that is between 11\% and 32\% higher than lexicon-based approaches, Word2Vec embeddings, and zero-shot classification with large language models such as GPT-4 for in-domain inference. Domain-adversarial training yields better out-of domain predictions than aggregate training while achieving comparable performance to zero-shot learning. 

MoralBERT holds substantial implications for future research. It opens up new possibilities for a more nuanced and context-sensitive understanding of moral narratives surrounding contentious social issues using NLP techniques. These insights can be instrumental in policy-making, social discourse, and conflict resolution by shedding light on the moral dimensions that underpin social stances.

\section{Related Work}
The Moral Foundations Dictionary (MFD)~\cite{Graham2009} was one of the first lexicons which was created to capture the moral rhetoric in text according to the initial five dimensions and their virtue/vice polarities defined by the MFT~\cite{Graham2011}.
The theory was subsequently extended with Liberty/Oppression as a foundation that deals with the domination and coercion by the more powerful upon the less powerful \cite{haidt2012righteous}. 
Until recently, much work on inferring morality from textual content does not include this foundation because the available linguistic resources are yet limited. 
Given its importance to recent controversial social discussions such as the vaccination debate,
Araque et al. \cite{araque2024novellexiconmoralfoundation} introduced the LibertyMFD lexicon generated based on aligned documents from online news sources with different worldviews. 

Traditionally, classification of moral elements in text has been approached via moral lexicons, lists of words depicting moral elements. Lexicons are generated manually \cite{Graham2009,Schwartz2012d}, via semi-automated methods \cite{Wilson2018,Araque2020}, or expanding a seed list with natural language processing (NLP) techniques \cite{Ponizovskiy2020,araque2022libertymfd}. The lexicons are then used to classify morality using text similarity \cite{Bahngat2020,Pavan2020MoralityText}. Moral elements have also been described as knowledge graphs to perform zero-shot classification \cite{asprino-etal-2022-uncovering}. More recent methods adopt instead supervised machine learning \cite{Qiu2022ValueNet:System,Alshomary2022,Liscio2022a,Huang2022LearningWeighting,Lan2022NewClassification}. Beiro et al. \cite{beiro2023moral} explored the role of Liberty/Oppresion in pro- and anti-vaccination Facebook posts using recurrent neural network classifiers with a short memory and entity linking information.

In general, there is a growing interest among researchers in analysing morality and the way modern machines perceive and capture it.  For instance, \citet{jiang2021can} presented DELPHI, an experimental framework based on neural networks capable of predicting judgements often aligned with human expectations (e.g., right and wrong). This work involved multiple experiments towards inclusive, ethically informed, and socially aware AI systems. Further, \citet{liu2022politics} presented POLITICS, a model for ideology prediction and stance detection. This model underwent training using novel ideology-focused pre-training objectives, which involved assessing articles on the same topic authored by media outlets with varying ideological perspectives. Another approach was introduced by \citet{mokhberian2020moral}  utilising unsupervised techniques to identify moral framing biases in news media.

More closely related to our work, Trager and colleagues \cite{trager2022moral} presented baseline models for moral values prediction using a pre-trained BERT model fine-tuned in Moral Foundation Reddit Corpus. However, this study is limited to a in-domain work and might not generalise well in other domains. 
Other studies have introduced out-of-domain approaches moral MFT predictions and presented techniques to enhance model generalisability \cite{nguyen2024measuring, guo2023data}.
\citet{guo2023data} presented a multi-label model for predicting moral values while using the domain adversarial training framework proposed by \citet{ganin2015unsupervised} to align multiple datasets. 
Commonly moral classification studies in textual content utilised BERT (Bidirectional Encoder Representations from Transformers) \cite{devlin2018bert}. Due to BERT's widespread adoption, several versions and successor models, including RoBERTa, T5, and DistilBERT, have been developed to effectively tackle a variety of tasks across multiple domains \cite{bird2023chatbot}. 

Recent studies have also explored the capabilities LLMs in understanding moral judgments. Ganguli et al. \cite{ganguli2023capacity} showed that LLMs trained with reinforcement learning from human feedback can morally self-correct to avoid harmful outputs. These models can follow instructions and learn complex normative concepts like stereotyping, bias, and discrimination. Zhou et al. \cite{zhou2023rethinking} proposed a theory-guided framework for prompting GPT-4 to perform moral reasoning based on established moral theories, demonstrating its capability to understand and make judgments according to these theories while aligning with human-annotated morality datasets. Scherrer et al. \cite{scherrer2024evaluating} assessed how LLMs encode moral beliefs, finding that in clear-cut scenarios, LLMs align with common sense, but in ambiguous situations, they often express uncertainty.
Although most recent LLMs have shown a great performance in understanding complex societal themes, researchers from different fields have shown that smaller but more specialised models like BERT can still reach better accuracy in supervised learning tasks \cite{hernandez2023we, chen2024evaluating}.

Building on previous studies, here we present in-domain and out-of domain moral foundations predictions from three major social media platforms. Unlike most of the studies, which only discuss 5 major MFT dimensions, we also analyse Liberty/Oppression foundation \cite{haidt2012righteous}. 
Moreover, we built an extensive set of experiments while comparing MoralBERT models with MoralStrength lexicon \cite{Araque2020} Word2Vec model with Random Forrest and zero-shot GPT-4.

\section{Data}

In this study, we employ three datasets sourced from major social media platforms, all manually annotated for their moral content according to the MFT.

First, we use the Moral Foundations Twitter Corpus (MFTC), a collection of seven distinct datasets totalling 35,108 tweets that have been hand-annotated by at least three trained annotators for five moral foundations (Liberty was not included, see below), each with vice/virtue polarities, resulting in a total of 10 labels \cite{hoover2020moral}. Also  a ``non-moral'' label was used for tweets that are neutral or do not reflect any moral trait. Each tweet can have one or multiple moral labels. 
Final labels were determined by considering 50\% agreement among the annotators. 
Here we employ six of the seven MFTC datasets, in total 20,628 tweets, focusing on the most populous topic collections, namely, \textbf{Hurricane Sandy}, \textbf{Baltimore Protest}, \textbf{All Lives Matter}, \textbf{Davidson Hate Speech}, the \textbf{2016 US Presidential Election}, and \textbf{Black Lives Matter} (BLM). 
 
For Liberty/Oppression we incorporate newly available annotations for the BLM and 2016 Election datasets, collected via the same procedure and annotation scheme as MFTC \cite{araque2024novellexiconmoralfoundation}.

We also use 13,995 Reddit posts from the Moral Foundations Reddit Corpus (MFRC) \cite{trager2022moral}.
MFRC is organised into three buckets: \textbf{US politics} with subreddits \textit{conservative}, \textit{antiwork}, and \textit{politics}; \textbf{French politics} with subreddits \textit{conservative}, \textit{europe}, \textit{geopolitics}, \textit{neoliberal} and \textit{worldviews}; \textbf{Everyday Morality} with subreddits like \textit{IAmTheAsshole}, \textit{conffession}, \textit{nostalgia} and \textit{relationship\_advice}. Similarly to MFTC, at least three trained annotators were used and a 50\% agreement threshold was maintained for final labels.
MFRC includes annotations for Proportionality and Equality, which we combine and label as Fairness.
Similarly to MFTC, MFRC does not include annotations for the moral foundation of Liberty. 
Unlike MFTC, MFRC does not account for the polarity of moral foundations. To address this, we used VADER sentiment scores as weights for vice/virtue per foundation \cite{hopp2021extended}.

Lastly, we use a dataset of 1,509 Facebook posts related to \textbf{pro- and anti-vaccination}, each hand-annotated by nine researchers familiar with MFT following a similar annotation scheme to MFTC \cite{beiro2023moral}. Annotation labels cover virtue and vice polarity for each MFT category, including Liberty/Oppression, and ``non-moral'' labels were also used to denote either moral neutrality or lack of any moral trait. 
Cohen's kappa between annotators was 0.32, indicating fair agreement, but also speaking to the difficulty of detecting morality in text also for human annotators. 

We consider each dataset as a distinct domain for morality learning.
Distinct social media platforms possess diverse linguistic and social structural environments, potentially leading to variations in moral language \cite{curiskis2020evaluation}. 
They also differ in audience interests, content presentation/style, and spreading patterns \cite{priya2019should, jaidka2018facebook}.
Accordingly, we hypothesised that the expression of moral values in tweets versus Reddit comments versus Facebook posts may vary across training data sourced from the corresponding corpora described above.

\begin{figure*}[ht!]
    \centering
    \includegraphics[width=.7\linewidth]{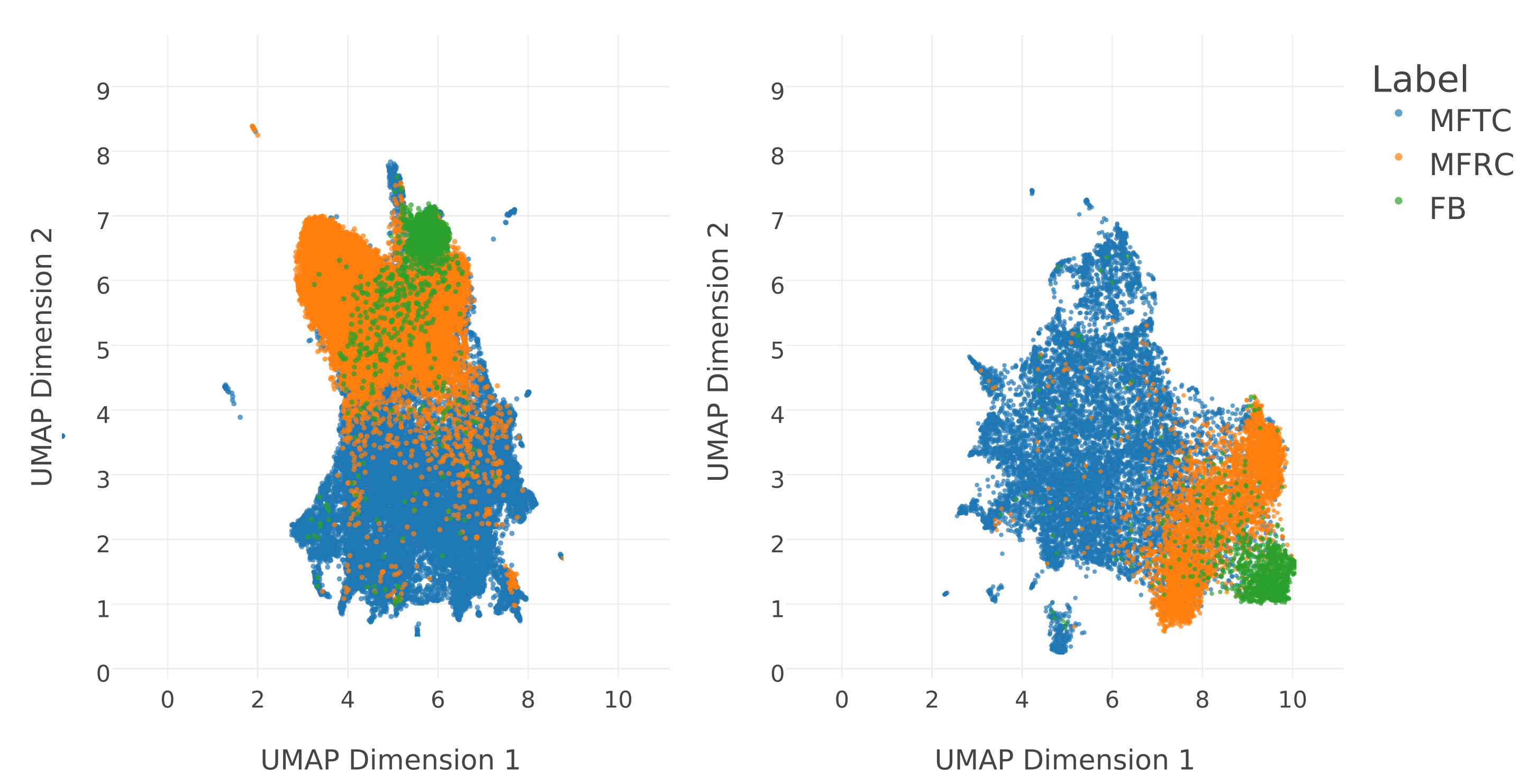}
    \Description{The figure shows the feature distribution of three Social Media domains. The left graph includes both moral and non-moral (neutral) labelled text, the right graph includes only moral labelled text. Dots represent mean-pooled BERT embeddings for each text example.}
    \caption{UMAP visualisation of feature distributions across datasets (domains). The left graph includes both moral and non-moral (neutral) labelled text, the right graph includes only moral labelled text. Dots represent mean-pooled BERT embeddings for each text example.}
    \label{fig:feature_embeddings}
\end{figure*}

Table~\ref{tab:Mft_text_distribution} illustrates the variation in the MFT label distribution including neutral (i.e., non-moral) text across the three social media datasets.
The left graph in Figure~\ref{fig:feature_embeddings} illustrates using manifold approximation and projection (UMAP \cite{mcinnes2018umap}) how the three corpora-domains differ in the feature embedding space. Feature distributions are generally distinct across the three datasets. There is some overlap between FB and MFRC, possibly because content in the respective platforms tends to be longer and more elaborate, while tweets are shorter due to character limits, resulting in more fragmented discussions and frequent updates.
This overlap becomes less prominent when neutral (non-moral) text is excluded (right graph in Figure~\ref{fig:feature_embeddings}), indicating that morally nuanced text is more clearly separated based on social media platform.

% \begin{comment}
% \end{comment}

\begin{table}[!ht]
\centering
\caption{Distribution of human-annotated moral values in the three social media corpora used in this study. \textsuperscript{\textdagger}Annotations for Liberty/Oppression are only available in FB (full corpus) and MFTC (BLM and 2016 US Election datasets).}
\setlength{\tabcolsep}{8pt}
\begin{tabular}{@{}lrrrr@{}} 
\toprule
 & MFTC & MFRC & FB & Total \\
\midrule
Care & 1658 & 737 & 357 & 2752 \\
Harm & 2027 & 1014 & 132 & 3173 \\
Fairness & 1575 & 623 & 173 & 2369 \\
Cheating & 2037 & 841 & 123 & 3001 \\
Loyalty & 1027 & 241 & 40 & 1308 \\
Betrayal & 1338 & 188 & 38 & 1564 \\
Authority & 824 & 330 & 110 & 1264 \\
Subversion & 565 & 357 & 204 & 1126 \\
Purity & 535 & 100 & 80 & 715 \\
Degradation & 746 & 187 & 112 & 1045 \\
Non-Moral & 7739 & 9842 & 367 & 17948\\
% \bottomrule
% \end{tabular}
% % \label{tab:Mft_text_distribution}
% \end{table}
% \begin{table}[!ht]
% \centering
% \caption{Distribution of Liberty/Oppression foundation in Social Media data.}
% \setlength{\tabcolsep}{8pt}
% % \begin{tabular}{lcccc}
% \toprule
\midrule
 % & \begin{tabular}[r]{@{}r@{}}MFTC \\ BLM\end{tabular} & \begin{tabular}[r]{@{}r@{}}MFTC \\ Election\end{tabular} &  &  \\
 % \midrule
Liberty\textsuperscript{\textdagger} & 2136 & 2284 & 140 & 4560 \\
Oppression\textsuperscript{\textdagger} & 1059 & 1028 & 65 & 2152 \\
Non-Moral\textsuperscript{\textdagger} & 692 & 735 & 367 & 1794 \\
\bottomrule
\end{tabular}
% \label{tab:liberty_oppression_distribution}
\label{tab:Mft_text_distribution}
\end{table}

\section{MoralBERT}

To capture moral expressions in social media discourse, we propose MoralBERT,\footnote{https://github.com/vjosapreniqi/MoralBERT} a series of transformer-based language models fine-tuned with the corpora presented in the previous section. These models use the BERT-base-uncased pretrained sequence classifier \cite{google-bert-base-uncased} with 768 hidden layers, 12 transformer layers and 110M parameters. We use the Adam optimiser with a learning rate of 5e-5 \cite{Devlin2019}. Due to data sparsity (see Table~\ref{tab:Mft_text_distribution}), we opted for a single-label classification approach, whereby each model predicts the presence or absence of a moral virtue or vice.

We further compare MoralBERT with MoralBERT$_{adv}$, an extension of the former with domain adversarial training  to account for heterogeneous training data. 
Following \citet{guo2023data}, we first obtain a domain invariant representation $h=W_{inv}e$ from the BERT embedding $e$, where
%\begin{equation*}
    % $h=W_{inv}e$
%\end{equation*}\\
% , and $W_{inv} \in \mathcal{R}^{768 \times 768}$
$W_{inv} \in \mathcal{R}^{768 \times 768}$ is a learnable matrix, which means that every element in this matrix is a parameter that can be adjusted during the training process. 
We then obtain predictions of moral values 
%\begin{equation*}
$\hat{y_m}=Softmax(W_1(ReLU(W_2h)))$
%\end{equation*}\\
where \( W_1 \in \mathbb{R}^{768 \times 768} \) 
and \( W_2 \in \mathbb{R}^{768 \times c} \)
are learnable matrices with parameters that the training process adjusts to optimize the model's performance. $c$ is the number of classes (0 represents neutral class, 1 represents the moral class), $ReLU$ is the rectified linear unit activation function and $Softmax$ is the normalised exponential function that gives the probability distribution over predicted classes. We include a domain classification head similar to the moral values classification head ($\hat{y_m}$) to obtain domain predictions ($\hat{y_d}$). The adversarial network connects the domain classifier head to the model via a gradient reversal layer, maximising domain classification loss while minimising the moral values prediction objective. We use the cross-entropy loss function for moral loss and domain loss representation. 

We also added two regularization terms \cite{guo2023data}, an L2 norm and a reconstruction loss, so that the original BERT does not get driven too far away from the original output embeddings by the fine-tuning process:
% \begin{align*}
%     L_{norm} &= ||W_{inv}-I||^2 \\
%     L_{rec} &= ||W_{rec}h-e||^2
% \end{align*}\\
%\vspace{1em} 
%\begin{center}
%\[
$L_{\text{norm}} = \|W_{\text{inv}} - I\|^2 $ 
and
%\]
%\[
$L_{\text{rec}} = \|W_{\text{rec}} h - e\|^2 $
%\]
%\end{center}
%\vspace{1em} 
where $I$ is the identity matrix and $W_{rec}$ 
%\in \mathcal{R}^{768 \times 768}$ 
a learnable matrix to reconstruct the original from the transformed embeddings.
We calculate the total loss by adding $L_{rec}$ and $L_{norm}$ to the moral values classification loss and the domain classification loss. 

For both MoralBERT and MoralBERT$_{adv}$ we assign class weights to address the class imbalance problem evident in Table~\ref{tab:Mft_text_distribution}. We do so by employing the approach of \citet{king2001logistic} so that for each class $c$ we compute 
% a weight as:
%\begin{equation*}
   $ weight_c=N / N_c$
%    \label{eqn:class_weighting}
%\end{equation*}\\
with $N$ being the total number of samples in the training data and $N_c$ the total number of samples in the training data belonging to class $c$.
We trained models with a batch size of 16, and input sequences are tokenised to a maximum length of 150 tokens determined by the maximum sentence size across the three combined datasets. Each model was trained for five epochs, and the model checkpoints from the best epoch were saved for testing.

\section{Experiments}
The performance of MoralBERT and MoralBERT$_{adv}$ was evaluated for in-domain and out-of-domain classification. To infer the 10 moral virtue/vice labels annotated across all three corpora (MFTC, MFRC, FB), in-domain models were trained using 80\% of the combined data from all datasets and tested in the remaining 20\%.
For out-of-domain models, we first train on two of the three corpora (e.g., MFTC and MFRC) and then test on the left-out dataset (e.g., FB). 
Due to the partial annotation of Liberty/Oppression in our datasets (MFTC BLM, MFTC Election, FB), we carried out separate experiments to infer this moral foundation following the same in-domain train-test split (80\%-20\%) and out-of-domain setup (training on FB and testing for MFTC, and vice versa). 
For all experiments we report the F1 Binary score, which focuses solely on moral labels and measures each model's accuracy in predicting true positives, and the F1 Macro score, which includes non-moral or neutral labels and measures each model's accuracy in predicting both true positives and true negatives.
In all data we use for fine-tuning and testing models, we cleaned the text by removing URLs, substituting mentions with "@user", removing hashtags, substituting emojis with their textual descriptions, and removing any non-ASCII characters using the \textit{re} Python library \cite{python-re}.

% \paragraph{Traditional Baseline Models} 
We employ two traditional baselines from previous works in this field. First, we use the MoralStrength lexicon \cite{Araque2020} 
% to make initial predictions for each category of moral values, which serves 
as a foundational estimate for each moral category. MoralStrength is an extension of MFD and offers a significantly larger set of morally annotated lemmas. It not only provides the moral valence score but also indicates the intensity of the lemma. Here, we use the MFT scores from MoralStrength to categorise values discretely to align with the output of MoralBert. 
Second, we quantise the textual data using Word2Vec, a widely used model in Natural Language Processing known for its word embedding capabilities \cite{mikolov2013distributed} and utilise a machine learning technique such as the Random Forest (RF) classifier with default parameters from the scikit-learn Python library \cite{pedregosa2011scikit} to predict each moral category. Word2Vec is a good method for handling large datasets and learning distributional properties of words, as well as syntactic and semantic word relationship \cite{mikolov2013distributed}. It demonstrates high performance in terms of both accuracy and computational efficiency \cite{gonzalez2023automatic}. However, Word2Vec embeddings do not model context, which makes them unsuitable for analyzing sentences.

We further compare MoralBERT and MoralBERT$_{adv}$ with a powerful LLM such as GPT-4 \cite{achiam2023gpt} deployed as a zero-shot classifier.   
LLMs like GPT-4 are trained on diverse text sources such as Wikipedia, GitHub, chat logs, books, and articles \cite{brown2020language}. This enables them to generalise and understand language across various domains \cite{doh2023lp}. The earlier model, GPT-3, contains 175 billion parameters, a figure vastly greater than BERT-base and BERT-large models (110M and 340M parameters) \cite{bosley2023we}. Given the size, cost, and significant energy consumption of these models, we used them for only 20\% of the data. The data selection was partially controlled. We selected 3,384 tweets from MFTC data, 2,793 Reddit posts from MFRC and 1509 posts from FB. We then prompted the classification task as follows:
% \begin{displayquote}
% \textbf{Prompt}: 

\textit{You will be provided with social media posts from Twitter, Reddit and Facebook, regarding different social topics. The social media posts will be delimited with \#\#\#\# characters. Classify each social media post into 12 Possible Moral Foundations as defined in Moral Foundation Theory. The available Moral Foundations are:} \{\texttt{Moral Foundations Tags}\}.
\textit{The explanation of the moral foundations is as follows:}  
\{\texttt{Description tags}\}.
\textit{This is a multi-label classification problem: where it's possible to assign one or multiple categories simultaneously. 
Report the results in JSON format such that the keys of the correct moral values are reported in a list}.
% \end{displayquote}

The \{\texttt{Moral Foundations Tags}\} 
represent the 12 moral virtues and vices, while the \{\texttt{Description Tags}\} provide a one-sentence description for each category.

\begin{table*}[!ht]
\centering
\small
\caption{In-domain prediction results for 10 Moral Foundations, showing F1 Binary and Macro average scores. Standard deviations are based on 1,000 bootstrap samples.
C. $=$ Care, H. $=$ Harm, Ch. $=$ Cheating, F. $=$ Fairness, L. $=$ Loyalty, B. $=$ Betrayal, A. $=$ Authority, S. $=$Subversion, P. $=$ Purity, D. $=$ Degradation.
}
\setlength{\tabcolsep}{4pt}
\begin{tabular}{@{}l|ccccc|ccccc@{}}
\toprule
 & \multicolumn{5}{c|}{\textbf{F1 Binary}} & \multicolumn{5}{c}{\textbf{F1 Macro}} \\
 \midrule
 & MS Lex. & W2V RF & GPT-4 & MoralBERT & MoralBERT$_{adv}$ & MS Lex. & W2V RF & GPT-4 & MoralBERT & MoralBERT$_{adv}$ \\
 \midrule
C. & .31 ± .02 & .14 ± .02 & .42 ± .01 & .48 ± .02 & \textbf{.50 ± .02} & .63 ± .01 & .55 ± .01 & .66 ± .01 & .71 ± .01 & \textbf{.73 ± .01} \\
H. & .38 ± .02 & .07 ± .01 & .41 ± .01 & .55 ± .01 & \textbf{.56 ± .02} & .65 ± .01 & .51 ± .01 & .64 ± .01 & .75 ± .01 & \textbf{.76 ± .01} \\
F. & .32 ± .02 & .35 ± .03 & .30 ± .01 & .56 ± .02 & \textbf{.57 ± .02} & .62 ± .01 & .66 ± .01 & .55 ± .01 & .76 ± .01 & \textbf{.77 ± .01} \\
Ch. & .19 ± .02 & .15 ± .02 & .34 ± .02 & .60 ± .01 & \textbf{.61 ± .01} & .57 ± .01 & .55 ± .01 & .64 ± .01 & .78 ± .01 & \textbf{.79 ± .01} \\
L. & .36 ± .02 & .28 ± .03 & .39 ± .02 & .57 ± .02 & \textbf{.64 ± .03} & .66 ± .01 & .63 ± .02 & .68 ± .01 & .78 ± .01 & \textbf{.81 ± .01} \\
B. & .14 ± .02 & .13 ± .02 & .18 ± .02 & .32 ± .03 & \textbf{.40 ± .04} & .55 ± .01 & .55 ± .01 & .57 ± .01 & .65 ± .02 & \textbf{.69 ± .02} \\
A. & .24 ± .02 & .22 ± .03 & .19 ± .01 & .37 ± .02 & \textbf{.39 ± .03} & .59 ± .01 & .60 ± .02 & .55 ± .01 & .67 ± .01 & \textbf{.69 ± .01} \\
S. & .25 ± .02 & .10 ± .03 & .22 ± .02 & .36 ± .02 & \textbf{.37 ± .02} & .60 ± .01 & .54 ± .01 & .58 ± .01 & \textbf{.67 ± .01} & \textbf{.67 ± .01} \\
P. & .17 ± .02 & .06 ± .03 & .44 ± .02 & .49 ± .03 & \textbf{.49 ± .02} & .56 ± .01 & .53 ± .01 & .71 ± .01 & \textbf{.74 ± .01} & .73 ± .01 \\
D. & \textbf{.28 ± .02} & .12 ± .03 & .21 ± .02 & .23 ± .02 & .25 ± .03 & \textbf{.62 ± .01} & .55 ± .01 & .59 ± .01 & .60 ± .01 & .61 ± .02 \\
\midrule
\textit{Avg}. & .26 ± .02 & .16 ± .03 & .31 ± .01 & .45 ± .02 & \textbf{.48 ± .02} & .61 ± .01 & .57 ± .01 & .62 ± .01 & .71 ± .01 & \textbf{.73 ± .01}
\\
\bottomrule
\end{tabular}
\label{tab:MFT_10_prediction_models}
\end{table*}

\begin{figure*}[ht!]
\centering
\begin{subfigure}[t]{1\linewidth}
    \includegraphics[width=\linewidth]{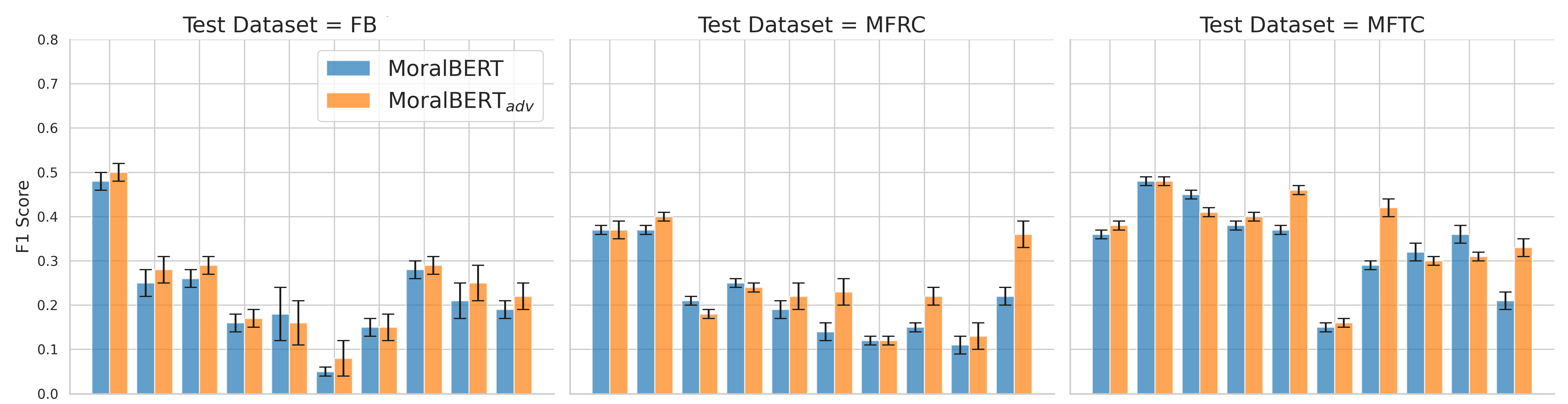}
    \caption{F1 Binary}
    \label{fig:F1_binary_out_of_domain}
\end{subfigure} \\
\begin{subfigure}[b]{1\linewidth}
    \includegraphics[width=\linewidth]{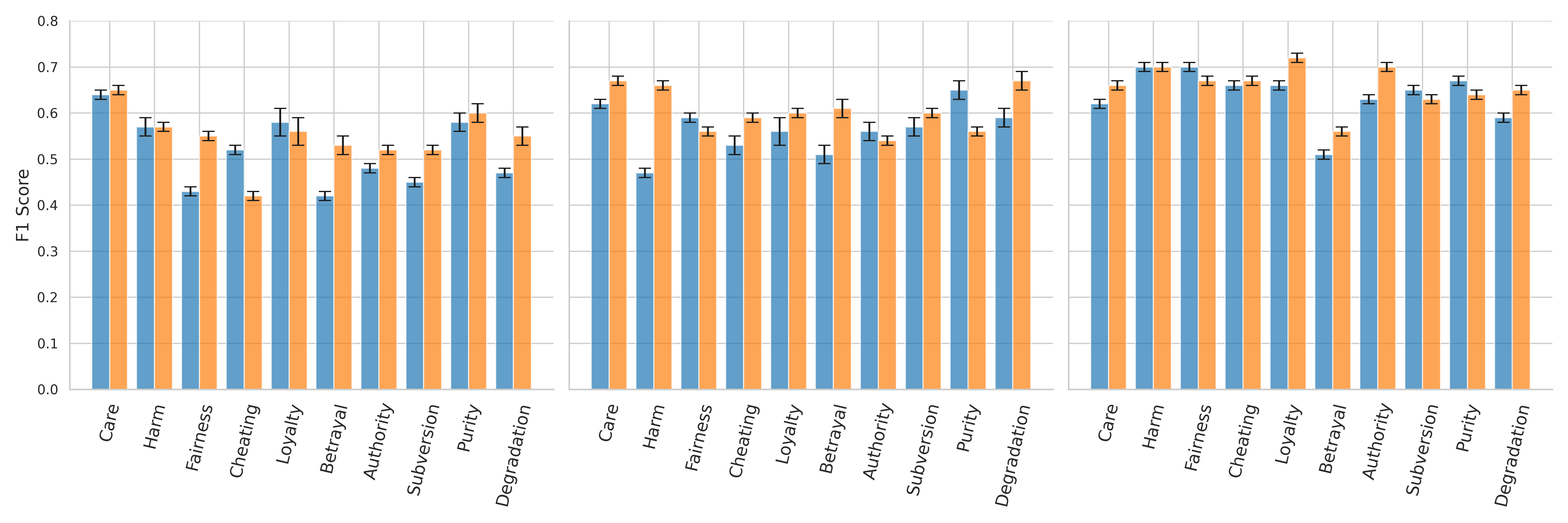}
    \caption{F1 Macro}
    \label{fig:F1_macro_out_of_domain}
\end{subfigure}
\caption{Out-of-domain classification: for each test dataset, models are fine-tuned on the other two datasets.  Bar heights represent F1 Binary and Macro average scores; error bars indicate standard deviation estimated via 1,000 bootstraps.}
\label{fig:F1_scores_out_of_domain}
\Description{The Figure represents Out-of-domain classification with MoralBERT and its adversarial version. For each test dataset, models are fine-tuned on the other two datasets.  Bar heights represent F1 Binary and Macro average scores; error bars indicate standard deviation estimated via 1,000 bootstraps.}
\end{figure*}

\begin{table*}[!ht]
    \centering
    \small
    \caption{
Zero-shot (GPT-4) versus out-of-domain (MoralBERT) classification, showing F1 Binary and Macro average scores and standard deviation estimated via 1,000 bootstraps. Models are fine-tuned on MFTC and MFRC and tested on FB.}
    \setlength{\tabcolsep}{10pt}
    \begin{tabular}{@{}l|ccc|ccc@{}}
    \toprule
     % & \multicolumn{6}{c}{Test dataset: FB Posts} \\
     % \midrule
     & \multicolumn{3}{c|}{\textbf{F1 Binary}} & \multicolumn{3}{c}{\textbf{F1 Macro}} \\
     \midrule
     & GPT-4 & MoralBERT & MoralBERT$_{adv}$ & GPT-4 & MoralBERT & MoralBERT$_{adv}$ \\
     \midrule
    Care & \textbf{.51 ± .02} & .48 ± .02 & .50 ± .02 & .62 ± .01 & .64 ± .01 & \textbf{.65 ± .01} \\
    Harm & .25 ± .02 & .25 ± .03 & \textbf{.28 ± .03} & .47 ± .01 & \textbf{.57 ± .02} & .57 ± .01 \\
    Fairness & \textbf{.34 ± .02} & .26 ± .02 & .29 ± .02 & \textbf{.59 ± .01} & .43 ± .01 & .55 ± .01 \\
    Cheating & .14 ± .03 & \textbf{.16 ± .02} & .17 ± .02 & \textbf{.53 ± .02} & .52 ± .01 & .42 ± .01 \\
    Loyalty & .14 ± .06 & \textbf{.18 ± .06} & .16 ± .05 & .56 ± .03 & \textbf{.58 ± .03} & .56 ± .03 \\
    Betrayal & .06 ± .03 & .05 ± .01 & \textbf{.08 ± .04} & .51 ± .02 & .42 ± .01 & \textbf{.53 ± .02} \\
    Authority & \textbf{.21 ± .03} & .15 ± .02 & .15 ± .03 & \textbf{.56 ± .02} & .48 ± .01 & .52 ± .01 \\
    Subversion & .23 ± .03 & .28 ± .02 & \textbf{.29 ± .02} & \textbf{.57 ± .02} & .45 ± .01 & .52 ± .01 \\
    Purity & \textbf{.34 ± .04} & .21 ± .04 & .25 ± .04 & \textbf{.65 ± .02} & .58 ± .02 & .60 ± .02 \\
    Degradation & \textbf{.25 ± .03} & .19 ± .02 & .22 ± .03 & \textbf{.59 ± .02} & .47 ± .01 & .55 ± .02 \\
    \midrule
    \textit{Avg.} & \textbf{.25 ± .03} & .22 ± .03 & .24 ± .03 & \textbf{.57 ± .02} & .51 ± .01 & .55 ± .02 \\
    \bottomrule
    \end{tabular}
    \label{tab:MFT_10_MoralBERT_GPT_FB_test}
\end{table*}

\begin{table*}[!ht]
\centering
\small
\caption{Examples of human-annotated and machine-learned moral values in social media discourse. GPT-4 is zero-shot classification; MoralBERT$_{adv}$ predictions are out-of-domain. R = Reddit; T = Twitter; F = Facebook.}
\begin{tabularx}{\linewidth}{
@{}>{\hsize=2.35\hsize\linewidth=\hsize}Y|
% >{\hsize=.5\hsize\linewidth=\hsize}Y|
>{\hsize=.55\hsize\linewidth=\hsize}Y|
>{\hsize=.55\hsize\linewidth=\hsize}Y|
>{\hsize=.55\hsize\linewidth=\hsize}Y@{}}
\toprule
\textbf{Text}   & \textbf{Human} & \textbf{GPT-4} & \textbf{MoralBERT}$_{adv}$ \\ 
\midrule
\textit{"And yet, more and more space and laws protect these people in their host countries. When are people in power going to wake up? OH RIGHT!, Le pen was, and it backfired on her"} [R] & Authority & Authority, Subversion & Care, Authority \\
\midrule
\textit{"I'll be blunt. I don't care whether a government (Macron's or anyone else's) has gender parity in its cabinet, all I actually hope is that the best people are chosen, regardless of having a wiener or not. For me this is what equality should be like"} [R] & Fairness & Fairness & Fairness \\
\midrule
\textit{"Those who deceive young men by selling war as an adventure are cruel monsters."} [T] & Harm, Cheating, Oppression & Care, Harm & Harm, Cheating, Betrayal \\
\midrule
\textit{"My tribute today to Sardar Patel-a Congress stalwart,who strove for communal harmony; dedicated his life to the unity"} [T] & Loyalty & Care, Fairness, Loyalty & Loyalty \\
\midrule
\textit{"Viruses and bacteria have no respect for religious beliefs. They will attack regardless. VERY few faiths promote an anti-vaccine agenda. Most consider the body to be a sacred gift that must receive proper care"}   [F] & Care, Purity & Care, Purity & Care, Purity \\ 
\midrule
\textit{"It's a travesty that kids are exposed to the insanity of Big Pharma. Parents must take the CO route and protect their kids. Meanwhile, get active in anti-vaccine groups since the PTB really do want mandatory vaccines or the kids will be given over to foster homes"} [F] & Care, Subversion, Liberty & Liberty, Oppression, Authority & Care, Subversion, Liberty, Oppression \\
\bottomrule
\end{tabularx}
\label{tab:qualitative_analysis}
\end{table*}

\begin{table*}[!ht]
\centering
\small
\caption{In-domain and out-of-domain predictions of Liberty/Oppression, showing F1 Binary and Macro average scores; standard deviation is calculated with 1000 bootstraps. }
% Set column separation
\setlength{\tabcolsep}{12pt}
\begin{tabular}{@{}l|ccc|ccc@{}}
\toprule
 & \multicolumn{3}{c|}{\textbf{F1 Binary}} & \multicolumn{3}{c}{\textbf{F1 Macro}} \\
 \midrule
 & \multicolumn{6}{c}{\textit{in-domain experiments}}\\
 \midrule
 & GPT-4 & MoralBERT & MoralBERT$_{adv}$ & GPT-4 & MoralBERT & MoralBERT$_{adv}$ \\
 \midrule
Liberty & .24 ± .02 & .63 ± .01 & \textbf{.66 ± .01} & .48 ± .01 & .70 ± .01 & \textbf{.71 ± .01} \\
Oppression & .17 ± .02 & \textbf{.45 ± .02} & .40 ± .02 & .51 ± .01 & \textbf{.68 ± .01} & .55 ± .01 \\
\midrule
 & \multicolumn{6}{c}{\textit{out-of-domain experiments, test dataset is FB (vaccination) }}\\
 \midrule

Liberty & \textbf{.39 ± .03} & .19 ± .02 & .19 ± .02 & \textbf{.62 ± .01} & .39 ± .01 & .39 ± .01 \\
Oppression & \textbf{.20 ± .03} & .05 ± .01 & .09 ± .01 & \textbf{.56 ± .02} & .49 ± .01 & .30 ± .01 \\
\midrule
 & \multicolumn{6}{c}{\textit{out-of-domain experiments, test dataset is MFTC (BLM and 2016 US Elections)}}
 \\
  \midrule
Liberty & .17 ± .01 & \textbf{.59 ± .01} & .57 ± .01  & .39 ± .01 & .52 ± .01 & \textbf{.53 ± .01} \\
Oppression  & .17 ± .02 & .25 ± .02 & \textbf{.27 ± .02} & .48 ± .01 & .49 ± .01 & \textbf{.50 ± .01} \\
\bottomrule
\end{tabular}
\label{tab:Liberty_Oppression_results}
\end{table*}

\section{Results}

Table \ref{tab:MFT_10_prediction_models} shows that MoralBERT$_{adv}$ had the highest performance for in-domain predictions. It achieved a 17\% higher F1 binary score compared to GPT-4, a 22\% higher score than MoralStrength, and a 32\% higher score Word2Vec with Random Forest models. The improved performance is also reflected in the F1 Macro scores. On average, MoralBERT$_{adv}$ surpasses GPT-4 by 11\%, MoralStrength by 12\%, and Word2Vec with Random Forest by 16\% in F1 macro score.

Figures \ref{fig:F1_binary_out_of_domain} and \ref{fig:F1_macro_out_of_domain} show that MoralBERT\(_{adv}\) performs marginally better than standard MoralBERT in F1 Binary and Macro average scores for out-of-domain predictions. 
For certain moral foundations, MoralBERT\(_{adv}\) shows significant improvements. For instance, Degradation predictions improve on MFRC and MFTC, and Loyalty and Authority predictions enhance on MFRC. As such, these moral foundations may be expressed differently across domains, and domain adaptation in MoralBERT$_{adv}$ enables the model to identify these patterns. 
 
We wanted to compare MoralBERT and GPT-4 for out-of-domain moral predictions. For this we used the MoralBERT model trained on MFTC and MFRC, and tested on FB, the smallest of our social media datasets with 1,509 posts, which allowed us to apply the zero-shot GPT-4 classification model to the entire Facebook dataset.
Inference on larger data could not be performed due to higher cost.
Table \ref{tab:MFT_10_MoralBERT_GPT_FB_test} reveals that the prediction results of MoralBERT, MoralBERT$_{adv}$, and GPT-4 are very similar, with GPT-4 achieving an average of 1\% higher F1 binary Score and 2\% higher F1 macro score. 

For Liberty/Oppression in-domain predictions showed in Figure \ref{tab:Liberty_Oppression_results}, the MoralBERT and MoralBERT$_{adv}$ performed better than GPT-4 with an average of 33\% higher F1 Binary Score and 19\% higher F1 Macro Score. 
In an out-of-domain setup for predicting this foundation in Facebook posts, zero-shot GPT-4 performed better than MoralBERT and MoralBERT\(_{adv}\), achieving an average F1 binary score 18\% higher and an F1 macro score 15\% higher.  The low performance of MoralBERT when tested on Facebook Data may be attributed to the marginal inner-annotator agreement (0.38 Cohen’s kappa coefficient) observed in the Facebook posts, indicating that these posts might be complex and ambiguous. In contrast, MoralBERT and MoralBERT$_{adv}$ performed significantly better when predicting Liberty/Oppression on MFTC tweets about BLM and the 2016 US Elections, with an average F1 binary score 25\% higher and an F1 macro score 8\% higher.

To quantitatively analyse the models' performance, we presented individual examples from the social media posts in the three datasets, annotated by human annotators, along with the results from the MoralBERT$_{adv}$ and the GPT-4 classification model in Table~\ref{tab:qualitative_analysis}. 
From the examples it can be seen that the text in the posts contains informal language, grammar mistakes, and many abbreviations. Further, some of the posts are written in an argumentative tone, and some use more personal and emotional nuances. We can also see that  Reddit comments and Facebook posts are typically much longer than tweets.
Since GPT-4 model is used as zero-shot approach it predicts moral labels based on the prompting request and the general knowledge that this model has Moral Foundation Theory. Previous works have shown that LLMs like GPT-4 can indeed perform moral reasoning through the lens of moral theories \cite{zhou2023rethinking} which is evident in our examples as well.
On the other hand, MoralBERT$_{adv}$ learns to align more closely with the trends seen in the fine-tuned annotation examples. However, MoralBERT$_{adv}$ often predicts both Liberty and Oppression even if only one is mentioned in the text.

\section{Discussion}
Moral values and judgments significantly influence our daily lives. Psychologists argue that moral judgment is not a rigorous reasoning process. Instead, it is influenced by more personal factors, including intuitions and emotions \cite{greene2002and}. 
Making moral judgments is intrinsically challenging, even for humans, due to lack of a universal standard \cite{zhou2023rethinking}. People from different beliefs and cultural backgrounds can have significantly different attitudes toward the same topic \cite{hu2021socioeconomic}. 
Furthermore, moral inferences are highly context-dependent \cite{ammanabrolu2022aligning} and different contexts can lead to distinct judgments \cite{zhou2023rethinking}. 
Similarly, Guo et al. \cite{guo2023data} showed that also the writing culture matters; in their study, using a model trained on MFTC to predict moral values in news articles (eMFD dataset \cite{hopp2021extended}) was shown significantly more challenging than predicting moral values on another Twitter dataset with discourse around COVID-19 vaccination. %\cite{rojecki2021moral}. 
In a more meticulous analysis, Lisco et al. \cite{liscio2023does} demonstrated that the predictability of moral values depends heavily on the distribution of moral rhetoric within a domain, and the further apart the domains are, the weaker the predictions of moral foundations become. 
This is evident in our study too, with MoralBERT models trained within the domain to be notably more successful at making moral inferences than those trained out-of-domain.
To mitigate this issue, we implemented the domain adversarial module (MoralBERT$_{adv}$ models) which resulted in marginal improvements in out-of-domain prediction models, demonstrating that out-of-domain moral inferences remain a challenging task. 

In our study, we face both challenges; diverse linguistic styles, since each dataset is sourced from a different social media platform, and a variety of social topics treated.
Another important factor in predicting moral values with language models like BERT is the distribution of moral labels in fine-tuning data \cite{guo2023data}. Unlike FB, the MFTC and MFRC corpora are highly imbalanced, with non-moral labelled text dominating over text labelled with moral values. The class weighting technique we employed played a significant role in addressing this issue in both the MoralBERT and MoralBERT$_{adv}$ models.

Regarding the experimental design, we opted for both a single-label approach, (predicting each moral value/vice separately), and a multi-label approach (predicting all moral values/vices at once) and as expected the results of the latter were significantly weaker. Thus we report the single-label experiments only.
The drop in the multi-label prediction approach relates to the intrinsic interdependence of moral dimensions \cite{Liscio2023}, which in our case is particularly challenging because some dimensions, like Care and Harm, often overlap with other moral dimensions. Instead, in the single-label setup, the prediction task is simpler, with the models focusing on learning specifically the dimension of interest and distinguishing it from morally neutral text. Furthermore, the single-label design approach in general showed better  performance for out of domain predictions, in line with the findings of \citet{guo2023data}.

In this work we introduce classification models for the Liberty/Oppression foundation, which were not previously considered in transformer-based approaches to morality inference from social media discourse \cite{nguyen2024measuring, guo2023data, trager2022moral}.
Given that Liberty/Oppression values are rooted in reason more than emotion \cite{iyer2012understanding}, they remain crucial for understanding decision-making across various contexts. They have been shown to be particularly relevant to current social issues such as the vaccination debate, poverty, and radicalisation \cite{mejova2023authority,araque2022libertymfd}.
In the present experiments, despite having significantly more limited training data resources, inference for Liberty/Oppression was comparable to that for the other 10 moral foundations. 

We benchmarked our MoralBERT and MoralBERT$_{adv}$ with both lexicon-based approaches, namely MoralStrength lexicon, Word2Vec, on top of Random Forrest as more traditional baselines, and large language models (LLM zero-shot GPT-4 classification). 
We showed that our models on average outperformed all other approaches for in-domain set-up with an approximate 11\% to 32\% improvement in F1 score, while for out-of domain the performances drops slightly as expected, but remain comparable with the GPT-4 classification results.
Overall, both MoralBERT and MoralBERT$_{adv}$ were better for predicting Liberty/Oppression in Twitter data.
Noteworthy is the fact that LLM models like GPT-4 are trained on billion of parameters, are extremely large, expensive, and consume significant amounts of energy, leading to various environmental implications and issues \cite{wu2022sustainable}.
This shows that BERT-based models can be just as effective as larger LLMs once  fine-tuned with considerably less resources. Moreover, the training is based on human annotators, ensuring that the models learn from  human moral reasoning. This is crucial especially for moral values assessment, misinterpretation of which can lead to social polarisation amplification.  
Another important point is that BERT-based approaches still provide interpretable results fundamental in assessing and examining how the model makes decisions on controversial social issues.

There is an ever-increasing interest in understanding moral values via natural language processing even in more artistic fields and beyond social media contexts. Recently, researchers have explored moral values in movie synopses \cite{gonzalez2023automatic} and lyrics \cite{preniqi2022more, preniqi2023soundscapes} using different dictionary and lexicon approaches. Our approach, utilising fine-tuned models for predicting moral values, will provide a valuable starting point for exploring morality in various contexts.
Understanding moral values from written content can greatly enhance communication and support social campaigns, but it also carries risks if used for malicious or manipulative purposes. Automatic annotation of morality in text can misrepresent individuals' moral positions or unfairly categorise them, leading to social stigmatisation and discrimination \cite{weidinger2021ethical}. 

Our research has certain limitations. First, although we gathered a substantial amount of textual content from three different social media platforms, a large portion of the data was labeled as non-moral or neutral. This leads to data imbalance issues which we tried to handle using a standard class weighting technique. Second, we could only partially use the data for the Liberty/Oppression foundation classification because this foundation was not present in all the datasets. Third, our study focused exclusively on English-language posts, which limits our understanding of how moral rhetoric is shaped across different cultures.

In the future, we aim to expand our investigations into multilingual models for understanding cross-cultural moral narratives.
Also, we will explore moral expressions in other domains, such as music lyrics, which often contain more complex linguistic structures and figurative expressions.
Additionally, we plan on employing techniques for distilling the knowledge \cite{hinton2015distilling} from LLMs like GPT-4 and LLama 2 \cite{touvron2023llama} in creating synthetic data which can be then used for fine-tuning language models like BERT models which are comparative for narrower tasks. This will help on further improving the current results in capturing moral values while reducing the need for manual annotation through the use of synthetic data generation \cite{ye-etal-2022-zerogen, meng2022generating}. By doing so, we can leverage the strengths of both types of models and improve our models' understanding of moral expressions in text across various domains and situations.
In general, we believe that this work is particularly timely, considering the current surge in research dedicated to identifying moral narratives in textual data. Even though there is room for improvement, our approach still holds significant value for the research community and beyond.

\begin{acks}
VP and IG are supported by PhD studentships from Queen Mary University of London's Centre for Doctoral Training in Data-informed Audience-centric Media Engineering. KK acknowledges support from the Lagrange Project of the Institute for Scientific Interchange Foundation (ISI Foundation) which is funded by Fondazione Cassa di Risparmio di Torino (Fondazione CRT). 
\end{acks}

% \bibliographystyle{ACM-Reference-Format}
% \bibliography{references}


\end{document}